%% file: PaperForReview.tex
\documentclass[10pt,twocolumn,letterpaper]{article}

\usepackage{cvpr}              

\usepackage{graphicx}
\usepackage{amsmath}
\usepackage{amssymb}
\usepackage{booktabs}
\usepackage{tabularx}
\usepackage{makecell}
\usepackage{multirow}
\usepackage{diagbox}
\usepackage{xcolor}

\usepackage[belowskip=-7pt,aboveskip=3pt]{caption}

\usepackage[pagebackref,breaklinks,colorlinks]{hyperref}

\usepackage[capitalize]{cleveref}
\crefname{section}{Sec.}{Secs.}
\Crefname{section}{Section}{Sections}
\Crefname{table}{Table}{Tables}
\crefname{table}{Tab.}{Tabs.}

\definecolor{roy}{rgb}{0.0, 0.0, 0.0}
\definecolor{howard}{rgb}{0.0, 0.0, 0.0}
\newcommand{\comment}[1]{\ignorespaces}
\newcommand{\roy}[1]{\textcolor{roy}{#1}}
\newcommand{\howard}[1]{\textcolor{howard}{#1}}

\def\cvprPaperID{1674} 
\def\confName{CVPR}
\def\confYear{2023}

\makeatletter
\def\thanks#1{\protected@xdef\@thanks{\@thanks
        \protect\footnotetext{#1}}}
\makeatother

\begin{document}

\title{Local Implicit Normalizing Flow for Arbitrary-Scale Image Super-Resolution}

\author{Jie-En Yao$^{*1}$, Li-Yuan Tsao$^{*1}$, Yi-Chen Lo$^{\dagger2}$, Roy Tseng$^{\dagger2}$, Chia-Che Chang$^{\dagger2}$, and Chun-Yi Lee$^{1}$\\
$^{1}$ElsaLab, National Tsing Hua University, $^{2}$MediaTek Inc.\\
{\tt\small \{{matt1129yao, lytsao}\}@gapp.nthu.edu.tw, \{{yichen.lo, roy.tseng, chia-che.chang}\}@mediatek.com}\\
{\tt\small cylee@cs.nthu.edu.tw}
\thanks{* and $\dagger$ indicate equal contribution.
 This work was developed during the internship of Jie-En Yao and Li-Yuan Tsao at MediaTek Inc.
}
}
\maketitle

\begin{abstract}
   \input{sections/abstract.tex}
\end{abstract}

\input{sections/all.tex}

{\small
\bibliographystyle{ieee_fullname}
\bibliography{egbib}
}

\end{document}

%% file: sections/abstract.tex
Flow-based methods have demonstrated promising results in addressing the ill-posed nature of super-resolution (SR) by learning the distribution of high-resolution (HR) images with the normalizing flow. However, these methods can only perform a predefined fixed-scale SR, limiting their potential in real-world applications. Meanwhile, arbitrary-scale SR has gained more attention and achieved great progress. Nonetheless, previous arbitrary-scale SR methods ignore the ill-posed problem and train the model with per-pixel L1 loss, leading to blurry SR outputs. In this work, we propose ``\textit{Local Implicit Normalizing Flow'' (LINF)} as a unified solution to the above problems. LINF models the distribution of texture details under different scaling factors with normalizing flow. Thus, LINF can generate photo-realistic HR images with rich texture details in arbitrary scale factors. We evaluate LINF with extensive experiments and show that LINF achieves the state-of-the-art perceptual quality compared with prior arbitrary-scale SR methods.

%% file: sections/all.tex
\vspace{-7pt}
\section{Introduction}
\label{sec::introduction}
\input{sections/introduction.tex}

\section{Related Work}
\label{sec::related_work}

\input{sections/related_work.tex}

\section{Methodology}
\label{sec::methodology}
\input{sections/methodology.tex}

\vspace{-5pt}
\section{Experimental Results}
\label{sec::experiments}
\input{sections/experiments.tex}

\section{Conclusion}
\label{sec::conclusions}
\input{sections/conclusions.tex}

\input{sections/acknowledgment.tex}

%% file: sections/introduction.tex
\howard{Arbitrary-scale image super-resolution (SR) has gained increasing attention recently due to its tremendous application potential.}
\howard{However, this field of study suffers from two major challenges. First, SR aims to reconstruct high-resolution (HR) image from a low-resolution (LR) counterpart by recovering the missing high-frequency information. This process is inherently ill-posed since the same LR image can yield many plausible HR solutions.}
\howard{Second, prior deep learning based SR approaches typically apply upsampling with a pre-defined scale in their network architectures, such as squeeze layer~\cite{srflow}, transposed convolution~\cite{fsrcnn}, and sub-pixel convolution~\cite{espcn}. Once the upsampling scale is determined, they are unable to further adjust the output resolutions without modifying their model architecture. This causes inflexibility in real-world applications.
As a result, discovering a way to perform arbitrary-scale SR and produce photo-realistic HR images from an LR image with a single model has become a crucial research direction.}
\comment{Single-image super-resolution (SR) aims to reconstruct the high-resolution (HR) image from a low-resolution (LR) counterpart by recovering the missing high-frequency information.}
\comment{There are two major challenges in SR. First, SR is an ill-posed problem by definition.}
\comment{Many plausible HR images can degrade to the same LR image.}
\comment{Second, in the real-world scenario, we often want the flexibility to upscale an LR image with an arbitrary scale.}
\comment{Hence, how to generate photo-realistic and arbitrary-scale HR images from one LR image has \comment{attracted}\roy{gained} increasing attention.}

\begin{figure}[t]
  \centering
  \includegraphics[width=0.95\linewidth]{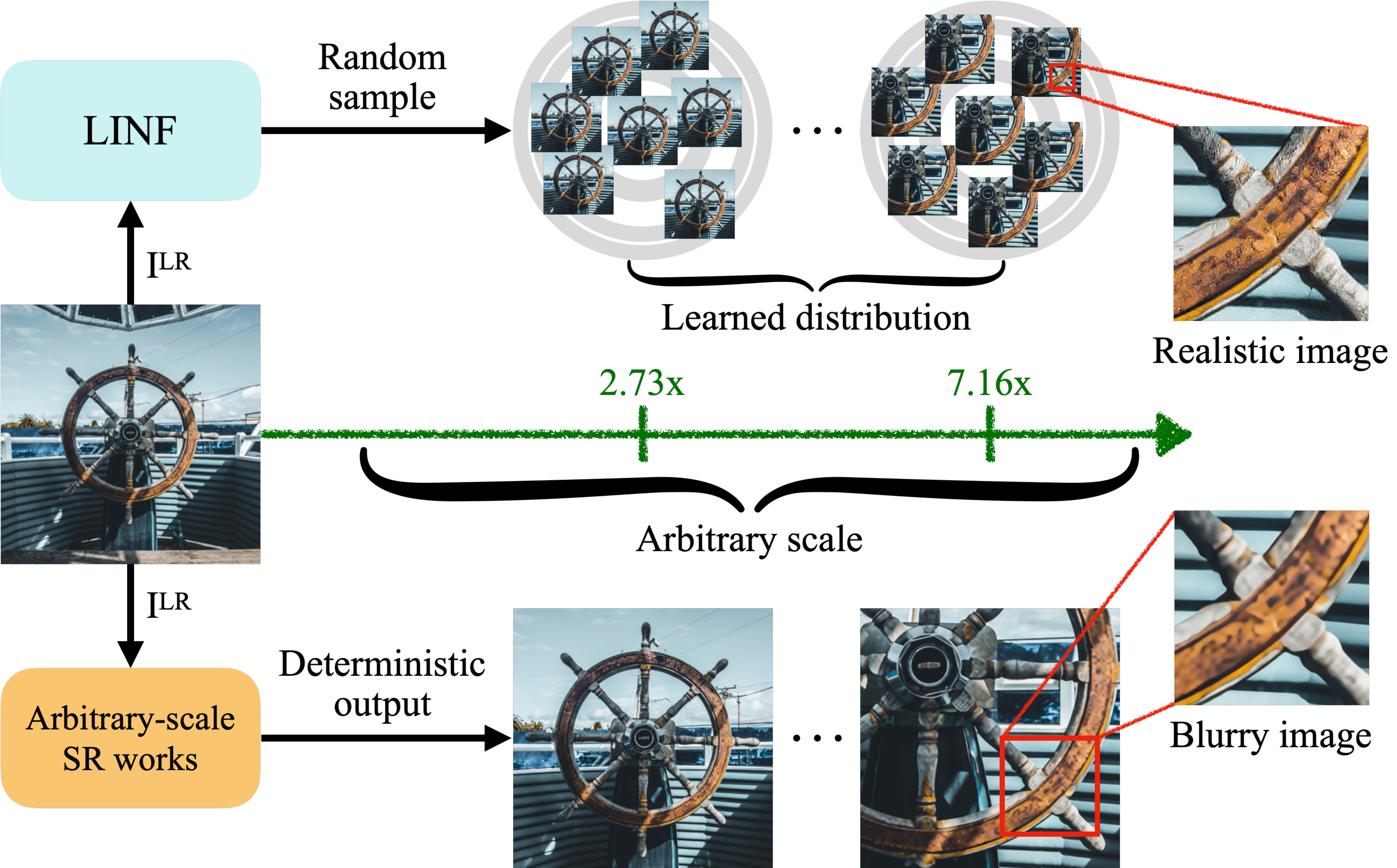}
  \caption{
    A comparison of the previous arbitrary-scale SR approaches and LINF. LINF models the distribution of texture details in HR images at arbitrary scales. Therefore, unlike the prior methods that tend to produce blurry images, LINF is able to generate arbitrary-scale HR images with rich and photo-realistic textures.
  }
  \label{fig:teaser}
\end{figure}

A natural approach to addressing the one-to-many inverse problem in SR is to consider the solution as a distribution. Consequently, a number of generative-based SR methods~\cite{esrgan, ranksrgan, srdiff, srflow, hcflow, larsr} have been proposed to tackle this ill-posed problem. 
Among them, flow-based SR methods show promise, as normalizing flow \cite{JimenezRezende2015VariationalIW, nice, realnvp, glow} offers several advantages over other generative models.
For instance, flow does not suffer from the training instability and mode collapse issues present in generative adversarial networks (GANs)~\cite{gan}.
Moreover, flow-based methods are computationally efficient compared to diffusion~\cite{ddpm} and autoregressive (AR)~\cite{pixelrnn, pixelcnn} models.
Representative flow-based models, such as SRFlow \cite{srflow} and HCFlow \cite{hcflow}, are able to generate high-quality SR images and achieve state-of-the-art results on the benchmarks. However, these methods are restricted to fixed-scale SR, limiting their applicability.


Another line of research focuses on arbitrary-scale SR. LIIF~\cite{liif} employs local implicit neural representation to represent images in a continuous domain. It achieves arbitrary-scale SR by replacing fixed-scale upsample modules with an MLP to query the pixel value at any coordinate. LTE~\cite{lte} further estimates the Fourier information at a given coordinate to make MLP focus on learning high-frequency details. However, these works did not explicitly account for the ill-posed nature of SR. They adopt a per-pixel $L1$ loss to train the model in a regression fashion. The reconstruction error favors the averaged output of all possible HR images, leading the model to generate blurry results.

Based on the observation above, combining flow-based SR model with the local implicit module is a promising direction \comment{where}\roy{in which} flow can account for the ill-posed nature of SR, and the local implicit module can serve as a solution to the arbitrary-scale challenge. Recently, LAR-SR~\cite{larsr} claimed that details in natural images are locally correlated without long-range dependency. Inspired by this insight, we formulated SR as a problem of learning the distribution of local texture patch. With the learned distribution, we perform super-resolution by \comment{separately}generating the local texture \roy{separately} for each non-overlapping patch in the HR image.

With the new problem formulation, we present Local Implicit Normalizing Flow (LINF) as the solution. Specifically, a coordinate conditional normalizing flow \comment{will} model\roy{s} \roy{the local texture patch distribution, which is conditioned on the LR image, the central coordinate of local patch, and the scaling factor.} \comment{the distribution of the local texture patch, where the distribution is conditioned on the LR image, the central coordinate of the local patch, and the scaling factor.} To provide the conditional signal for the flow model, we use the local implicit module to estimate Fourier information at each local patch. LINF excels the previous flow-based SR methods with the capability to upscale images with arbitrary scale factors. Different from prior arbitrary-scale SR methods, LINF explicitly addresses the ill-posed issue by learning the distribution of local texture patch. As shown in Fig~\ref{fig:teaser}, hence, LINF can generate HR images with rich and reasonable details instead of the over-smoothed ones.
Furthermore, \comment{LINF addresses the issue that generative models sometimes generate unpleasant artifacts by controlling the sampling temperature.}\roy{LINF can address the issue of unpleasant generative artifacts, a common drawback of generative models, by controlling the sampling temperature.} Specifically, the sampling temperature in normalizing flow controls the trade-off between PSNR (fidelity-oriented metric) and LPIPS~\cite{lpips} (perceptual-oriented metric). 
The contributions of this work can be summarized as follows:
\begin{itemize}
\item We proposed a novel LINF framework that leverages the advantages of a local implicit module and normalizing flow. 
To the best of our knowledge, LINF is the first framework that employs normalizing flow to generate photo-realistic HR images at arbitrary scales.
\item We validate the effectiveness of LINF to serve as a unified solution for the ill-posed and arbitrary-scale challenges in SR via quantitative and qualitative evidences. 
\item We examine the trade-offs between the fidelity- and perceptual-oriented metrics, and show that LINF does yield a better trade-off than the prior SR approaches.
\end{itemize}

%% file: sections/related_work.tex
In this section, we briefly review the previous deep learning based fixed-scale and arbitrary-scale SR methodologies.

\subsection{Fixed-Scale Super-Resolution} 
\label{subsec:SR}
A number of previous approaches have been proposed in the literature with an aim to learn mapping functions from given LR images to fixed-scale HR ones. These approaches can be broadly categorized into PSNR-oriented methods~\cite{srcnn, fsrcnn, espcn, edsr, rdn, swinir} and generative model based methods~\cite{srgan, sftgan, esrgan, ranksrgan, sr3, srdiff, cnf, srflow, adflow, hcflow, larsr}. 
The former category deterministically maps an LR image to an HR one using the standard L1 or L2 losses as the learning objectives. Despite the promising performance on the PSNR metric, the L1 or L2 losses adopted by such methods usually drives the models to predict the average of all plausible HR images~\cite{Bruna2016SuperResolutionWD, vggloss, srgan, srflow}, leading to an over-smoothed one. On the other hand, the latter category seeks to address the ill-posed nature of the SR problem by learning the distribution of possible HR images. Such methods include GAN-based SR, diffusion-based SR, flow-based SR, and AR-based SR. GAN-based SR methods~\cite{srgan, sftgan, esrgan, ranksrgan} train their SR models with adversarial loss, and are able to generate sharp and natural SR images. 
However, they sometimes suffer from training instability, mode collapse, and over-sharpen artifacts. 
Diffusion-based SR methods~\cite{srdiff, sr3} generate an HR image by iteratively refining a Gaussian noise using a denoising model conditioned on the corresponding LR image. These methods are promising and effective, nevertheless, the slow iterative denoise processes limit their practical applications. 
Flow-based SR methods~\cite{cnf, srflow, adflow, hcflow} utilize invertible normalizing flow models to parameterize a distribution. They are promising and achieve state-of-the-art results on the benchmark as they possess several advantages over other generative models, as discussed in Section~\ref{sec::introduction}.
Among these methods, SRFlow~\cite{srflow} first pioneered the flow-based SR domain. It was then followed by HCFlow~\cite{hcflow}, which designed a hierarchical conditional mechanism in the flow framework and achieved better performance than SRFlow.  Recently, LAR-SR~\cite{larsr} introduced the first AR-based SR model. It divides an image into non-overlapping patches, and learns to generate local textures in these patches using a local autoregressive model.

\subsection{Arbitrary-Scale Super-Resolution} 
\label{subsec:Arbi_SR}

Despite the successes, the approaches discussed in Section~\ref{subsec:SR} are only able to super-resolve LR images with predefined upsampling scales, which are usually restricted to certain integer values~(e.g., $2\times$$\sim$$4\times$).   Meta-SR~\cite{metasr} first attempted to address this limitation by introducing a meta-learning based method to adaptively predict the weights of the upscaling filters for each scaling factor. This avenue is then explored by a number of follow-up endeavors~\cite{rsn, arbsr, liif, ultrasr, ipe, itsrn, lte, itsrn++}.  RSAN~\cite{rsn} proposed a scale attention module to learn informative features according to the specified scaling factor. ArbSR~\cite{arbsr} employed a plug-in module to perform scale-aware feature adaptation and scale-aware upsampling.  Recently, LIIF~\cite{liif} introduced the concept of local implicit neural representation.
Given necessary feature embeddings and a coordinate in the real coordinate space $\mathbb{R}^2$, LIIF enables the RGB value of that pixel coordinate to be decoded by a multilayer perceptron (MLP). Inspired by~\cite{spectralbias, siren, nerf, fourierfeature}, UltraSR~\cite{ultrasr} and IPE~\cite{ipe} enhanced LIIF by introducing positional encoding to the framework, allowing it to focus more on high-frequency details. The authors of LTE~\cite{lte} further introduced the use of Fourier features in their local texture estimator for estimating the dominant frequencies of an image.

%% file: sections/methodology.tex
In this section, we first formally define the SR problem concerned by this paper, and provide an overview of the proposed framework. Then, we elaborate on the details of its modules, followed by a discussion of our training scheme.

\begin{figure}[t]
  \centering
  \includegraphics[width=0.95\linewidth]{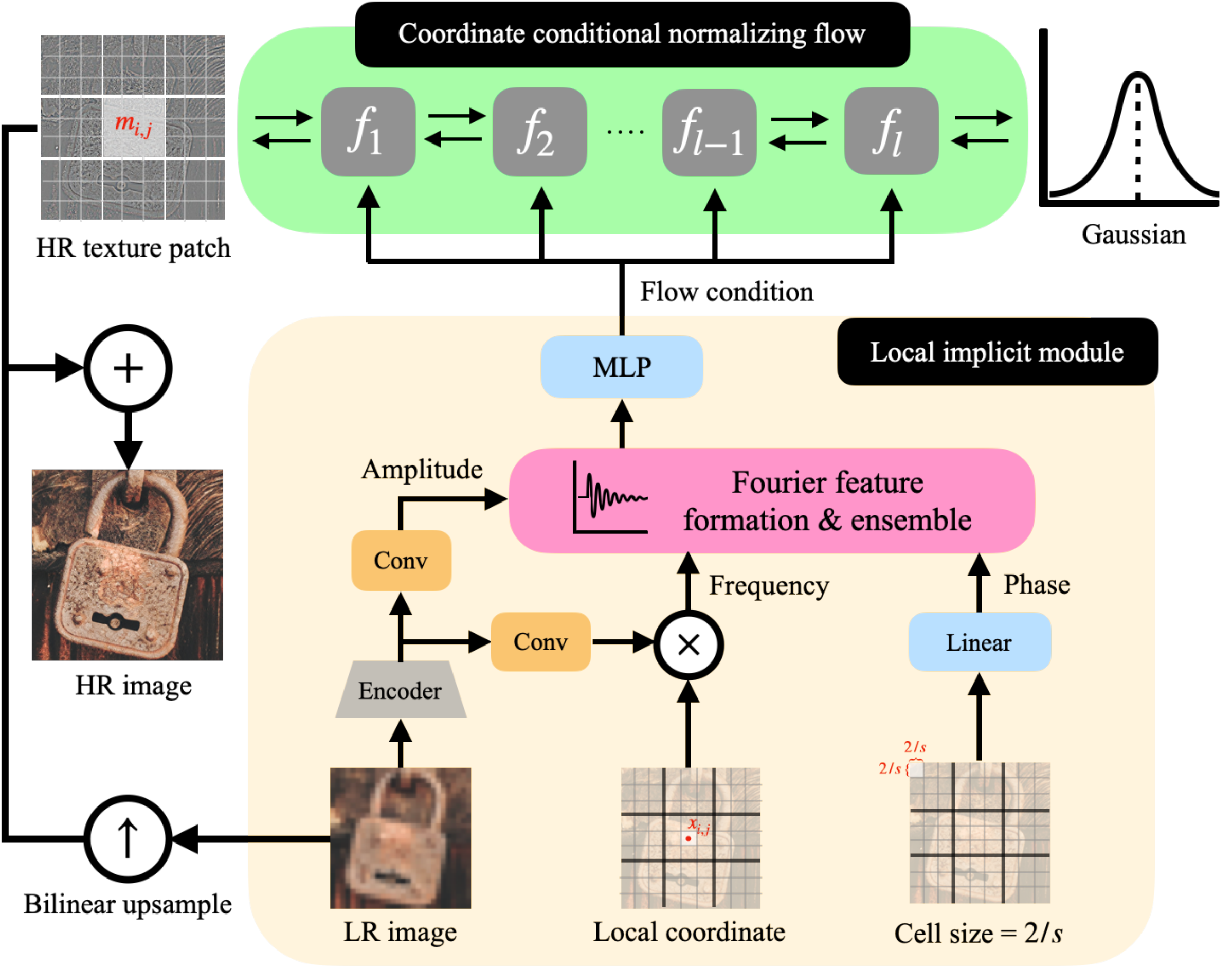}
  \caption{
    An illustration of the proposed LINF framework.  LINF consists of two parts. The local implicit model first encodes an LR image, a local coordinate and a cell into Fourier features, which is followed by an MLP for generating the conditional parameters. The flow model then leverages these parameters to learn a bijective mapping between a local texture patch space and a latent space. 
  }
  \label{fig:overview}
\end{figure}


\vspace{-7pt}
\paragraph{Problem definition.}

Given an LR image \comment{denoted as} $I^{LR} \in \mathbb{R}^{H \times W \times 3}$ and an arbitrary scaling factor $s$, the objective of this work is to generate an HR image \comment{denoted as} $I^{HR} \in \mathbb{R}^{sH \times sW \times 3}$, where $H$ and $W$ represent the height and width of the LR image.  Different from previous works, we formulate SR as a problem of learning the distributions of \textit{local texture patches} by normalizing flow, where `\textit{texture}' is defined as the residual between an HR image and \comment{the one generated by bilinear interpolation.}\roy{the bilinearly upsampled LR counterpart.} These local texture patches are constructed by grouping $sH \times sW$ pixels of $I^{HR}$ into $h \times w$ non-overlapping patches of size $n \times n$ pixels, where  $h = \lceil sH / n \rceil, w = \lceil sW / n \rceil$. The target distribution of a local texture patch $m_{i, j}$ to be learned can be formulated as a conditional probability distribution $p(m_{i, j}|I^{LR},x_{i, j},s)$, where $(i, j)$ represent the patch index, and $x_{i, j} \in \mathbb{R}^{2}$ denotes the center coordinate of $m_{i, j}$. The predicted local texture patches are aggregated together to form $I^{HR}_{texture}\in \mathbb{R}^{sH \times sW \times 3}$, which is then combined with a bilinearly upsampled image $\roy{{I^{LR}}_{\uparrow}}\in \mathbb{R}^{sH \times sW \times 3}$ via element-wise addition to derive the final HR image $I^{HR}$.




\vspace{-6pt}
\paragraph{Overview.}
Fig.~\ref{fig:overview} provides an overview of the LINF framework, which consists of two modules: (1) a local implicit module, and (2) a coordinate conditional normalizing flow (or simply ``\textit{the flow model}" hereafter). The former generates the conditional parameters for the latter, enabling LINF to take advantages of both local implicit neural representation and normalizing flow.  Specifically, the former first derives the local Fourier features~\cite{lte} from $I^{LR}$, $x_{i, j}$, and $s$. The proposed Fourier feature ensemble is then applied on the extracted features.
Finally, \roy{given the ensembled feature,} the latter utilizes an MLP to generate the parameters for the flow model to approximate $p(m_{i, j}|I^{LR},x_{i, j},s)$. We next elaborate on their details and the training strategy.



\subsection{Coordinate Conditional Normalizing Flow} 
\label{subsec:CCNF}

Normalizing flow approximates a target distribution by learning a bijective mapping $\boldsymbol{f}_\theta = f_1 \circ f_2 \circ ... \circ f_l$ between a target space and a latent space, where $\boldsymbol{f}_\theta$ denotes a flow model parameterized by $\theta$, and $f_1$ to $f_l$ represent $l$ invertible flow layers. In LINF, the flow model approximates such a mapping between a local texture patch distribution $p(m_{i, j}|I^{LR},x_{i, j},s)$ and a Gaussian distribution $p_z(z)$ as:
\begin{equation} \label{eq:1}
    m_{i, j}=h_0 \overset{f_1}{\underset{f^{-1}_1}\rightleftarrows} h_1 \overset{f_2}{\underset{f^{-1}_2}\rightleftarrows} ...~h_{k-1}
    \overset{f_k}{\underset{f^{-1}_k}\rightleftarrows} h_{k}~...
    \overset{f_l}{\underset{f^{-1}_l}\rightleftarrows} h_l=z, 
\end{equation}
where $z\sim\mathcal{N}(0, \tau)$ is a Gaussian random variable, $\tau$ is a temperature coefficient,
 $h_k=f_{k}(h_{k-1})$, $k\in[1,...,l]$, denotes a latent variable in the transformation process, and $f^{-1}_k$ is the inverse of $f_k$. By applying the change of variable technique, the mapping of the two distributions $p(m_{i, j}|I^{LR},x_{i, j},s)$ and $p_z(z)$ can be expressed as follows:
\begin{equation} \label{eq:2}
\begin{aligned}
    log\ p_\theta(m_{i, j}|I^{LR},x_{i, j},s) &= log\ p_z(z)\ \\
    &+\sum_{k=1}^{l}log\ \Big|det\frac{\partial f_k(h_{k-1})}{\partial h_{k-1}}\Big|.
\end{aligned}
\end{equation}
The term $log\ |det\frac{\partial f_k(h_{k-1})}{\partial h_{k-1}}|$ is the logarithm of the absolute Jacobian determinant of $f_k$. As $I^{HR}_{texture}$ (and hence, the local texture patches) can be directly derived from $I^{HR}$, $I^{LR}$, and $s$ during the training phase, the flow model can be optimized by minimizing the negative log-likelihood loss.  
During the inference phase, the flow model is used to infer local texture patches by transforming sampled $z$'s with $f^{-1}$. Note that the values of $\tau$ are different during the training and the inference phases, which are discussed in Section~\ref{sec::experiments}.
\vspace{-6pt}
\paragraph{Implementation details.}

Since the objective of our flow model is to approximate the distributions of local texture patches rather than an entire image, it is implemented with a relatively straightforward model architecture. The flow model is composed of ten flow layers, each of which consists of a linear layer and an affine injector layer proposed in~\cite{srflow}. Each linear layer $k$ is parameterized by a learnable pair of weight matrix $\mathcal{W}_k$ and bias $\beta_k$. The forward and inverse operations of the linear layer can be formulated as:
\begin{equation} \label{eq:3}
    h_{k} = \mathcal{W}_k h_{k-1} + \beta_k\ \ ,\ \ h_{k-1} = \mathcal{W}_k^{-1} (h_{k} - \beta_k),
\end{equation}
where $\mathcal{W}_k^{-1}$ is the inverse matrix of $\mathcal{W}_k$. The Jacobian determinant of a linear layer is simply the determinant of the weight matrix $\mathcal{W}_k$. Since the dimension of a local texture patch is relatively small (i.e., $n \times n$ pixels), calculating the inverse and determinant of the weight matrix $\mathcal{W}_k$ is feasible.

On the other hand, the affine injector layers are employed to enable two conditional parameters $\alpha$ and $\phi$ generated from the local implicit module to be fed into the flow model. The incorporation of these layers allows the distribution of a local texture patch $m_{i, j}$ to be conditioned on $I^{LR}$, $x_{i, j}$, and $s$. The conditional parameters are utilized to perform element-wise shifting and scaling of latent $h$, expressed as:
\begin{equation} \label{eq:4}
    h_{k} = \alpha_k \odot h_{k-1} + \phi_k\ \ ,\ \ h_{k-1} = (h_{k} - \phi_k) / \alpha_k,
\end{equation}
where $k$ denotes the index of a certain affine injector layer, and $\odot$ represents element-wise multiplication.  The log-determinant of an affine injector layer is computed as $\sum log(\alpha_k)$, which sums over all dimensions of indices~\cite{srflow}.



\subsection{Local Implicit Module} 
\label{subsec:LINR}

The goal of the local implicit module is to generate  conditional parameters $\alpha$ and $\phi$ from the local Fourier features extracted from $I^{LR}$, $x_q$, and $s$. This can be formulated as:
\begin{equation} \label{eq:6}
    \alpha, \phi = g_{\Phi}(E_{\Psi}(v^*, x_q-x^*, c)),
\end{equation}
where $g_{\Phi}$ represents the parameter generation function implemented as an MLP, $x_q$ is the center coordinate of a queried local texture patch in $I^{HR}$, $v^*$ is the feature vector of the 2D LR coordinate $x^*$ which is nearest to $x_q$ in the continuous image domain~\cite{liif}, $c=2/s$ denotes the cell size, and $x_q - x^*$ is known as the relative coordinate.  Following\cite{lte}, the local implicit module employs a local texture estimator $E_\Psi$ to extract the Fourier features given any arbitrary $x_q$. This function can be expressed as follows:
\begin{equation} \label{eq:5}
    E_{\Psi}(v^*, x_q-x^*, c): A \odot \begin{bmatrix} cos(\pi F (x_q-x^*) + P) \\sin(\pi F (x_q-x^*) + P) \end{bmatrix},
\end{equation}
where $\odot$ denotes element-wise multiplication, and $A$, $F$, $P$ are the Fourier features extracted by three distinct functions:
\begin{equation} \label{eq:Fourier}
    A=E_a(v^*), F=E_f(v^*), P=E_p(c),
\end{equation}
where $E_a$, $E_f$, and $E_p$ are the functions for estimating amplitudes, frequencies, and phases, respectively. In this work, the former two are implemented with convolutional layers, while the latter is implemented as an MLP. Given the number of frequencies to be modeled as $K$, the dimensions of these features are $A \in \mathbb{R}^{2K}$, $F \in \mathbb{R}^{K \times 2}$, and $P\in \mathbb{R}^{K}$.

\vspace{-6pt}
\paragraph{Fourier feature ensemble.}

To avoid color discontinuity when two adjacent pixels select two different feature vectors, a local ensemble method was proposed
in~\cite{liif} to allow RGB values to be queried from the nearest four feature vectors around $x_q$ and \comment{fused} \roy{fuse them} with bilinear interpolation. If this method is employed, the forward and inverse transformation of our flow model $f_\theta$ would be expressed as follows:
\begin{equation} \label{eq:7}
\begin{aligned}
    z = \sum_{j\in\Upsilon}w_j*f_{\theta}(patch;g_{\Phi}(E_{\Psi}(v_j, x_q-x_j, c))) \\
    patch = \sum_{j\in\Upsilon}w_j*f^{-1}_{\theta}(z;g_{\Phi}(E_{\Psi}(v_j, x_q-x_j, c))),
\end{aligned}    
\end{equation}
where $\Upsilon$ is the set of four nearest feature vectors, and $w_j$ is the derived weight for performing bilinear interpolation. 

Albeit effective, local ensemble requires four forward passes of the local texture estimator $E_{\Psi}$, the parameter generator $g_{\Phi}$, and the flow model $f_\theta$. To deal with this drawback, our local implicit module employs a different approach named ``\textit{Fourier feature ensemble}'' to streamline the computation. Instead of directly generating four RGB samples and then fuse them in the image domain, we propose to ensemble the  four nearest feature vectors right after the local texture estimator $E_{\Psi}$. More specifically, these feature vectors are concatenated to form an ensemble $\kappa = concat(\{w_j * E_{\Psi}(v_j, x_q-x_j, c), \forall j\in\Upsilon\})$, in which each feature vector is weighted by $w_j$ to allow the model to focus more on closer feature vectors. The proposed technique \comment{enables} \roy{requires} $g_{\Phi}$ and $f_\theta$ to perform only one forward pass to capture the same amount of information as the local ensemble method and deliver same performance. It is expressed as:
\begin{equation} \label{eq:8}
\begin{aligned}
    z = f_{\theta}(patch;g_{\Phi}(\kappa)); 
    patch = f^{-1}_{\theta}(z;g_{\Phi}(\kappa)).
\end{aligned}  
\end{equation}

\subsection{Training Scheme} 

LINF employs a two-stage training scheme. In the first stage, it is trained only with the negative log-likelihood loss $L_{nll}$. In the second stage, it is fine-tuned with an additional L1 loss on predicted pixels $L_{pixel}$, and the VGG perceptual loss~\cite{vggloss} \roy{on the patches predicted by the flow model} $L_{vgg}$.
The total loss function $L$ can be formulated as follows:
\begin{equation} \label{eq:9}
\begin{aligned}
    L = &\lambda_1 L_{nll}(patch_{gt}) + \lambda_2 L_{pixel}(patch_{gt}, patch_{\tau=0}) \\
    & + \lambda_3 L_{vgg}(patch_{gt}, patch_{\tau=0.8}),
\end{aligned}  
\end{equation}
where $\lambda_1$ $\lambda_2$, and $\lambda_3$ are the scaling parameters, $patch_{gt}$ denotes the ground-truth local texture patch, and ($patch_{\tau=0}$, $patch_{\tau=0.8}$) represent the local texture patches predicted by LINF with temperature $\tau=0$ and $\tau=0.8$, respectively.


%% file: sections/experiments.tex
In this section, we report the experimental results, present the ablation analyses, and discuss the implications.

\subsection{Experimental Setups} 
\label{subsec:setup}
In this section, we describe the experimental setups. We compare LINF with previous arbitrary-scale SR methods and generative SR models to show that LINF is able to generate photo-realistic HR images for arbitrary scaling factors.

\vspace{-5pt}
\paragraph{Arbitrary-scale SR.}
We use the DIV2K\cite{div2k} dataset for training 
and evaluate the performance on several widely used SR benchmark datasets, including Set5\cite{set5}, Set14\cite{set14}, B100\cite{b100}, and Urban100\cite{urban100}. 
To compare our LINF with the prior pixel-wise SR 
methods\cite{liif, lte},
we set the patch size $n$ to $1 \times 1$, which models
the distribution of a single pixel. 
We use three different encoders, EDSR-baseline\cite{edsr}, RDN\cite{rdn}, and SwinIR\cite{swinir},
to extract features of LR images. In the first training stage, we train the models for $1,000$ epochs, with a learning rate of
$1 \times 10^{-4}$, which is
halved at epochs [200,\,400,\,600,\,800] for EDSR-baseline and RDN, and at epochs [500,\,800,\,900,\,950] for SwinIR. 
In the second stage, we fine-tune EDSR-baseline and RDN for $1,000$ epochs, and SwinIR for $1,500$ epochs, with a fine-tune
learning rate of $5 \times 10^{-5}$, which is
halved at epochs [200,\,400,\,600,\,800] for EDSR-baseline and RDN, and at epochs [800,\,1100,\,1300,\,1400] for SwinIR. The parameters in Eq.~(\ref{eq:9}) are set by $\lambda_1 = 5 \times 10^{-4}$, $\lambda_2 = 1$, and $\lambda_3 = 0$. 
The Adam optimizer is used for training. The batch size is $16$ for EDSR-baseline and RDN, and $32$ for SwinIR.

\vspace{-5pt}
\paragraph{Generative SR.}

For generative SR, our models are trained on both the DIV2K\cite{div2k} and Flickr2K\cite{flickr2k} datasets, with performance evaluation conducted using
the DIV2K validation set. 
To effectively capture the underlying texture distribution, we set the patch size $n$ to $3 \times 3$.
The RRDB architecture~\cite{esrgan} is employed as the encoder. The training parameters, such as epoch, learning rate, batch size, and optimizer settings, are maintained in alignment with RDN.
Moreover, we set the loss weighting parameters to be $\lambda_1 = 5 \times 10^{-4}$, $\lambda_2 = 1$, and $\lambda_3 = 2.5 \times 10^{-2}$, respectively.

\vspace{-5pt}
\paragraph{Training strategy.}

In the proposed LINF methodology, the model is trained utilizing scaling factors within a continuous range from $\times1$ to $\times4$.
In practice, for each data sample within a mini-batch, a scale denoted as $s$ is obtained by sampling from a uniform distribution $U(1, 4)$.
The LR image dimensions are set to $48 \times 48$ pixels. As a result, this configuration necessitates the cropping of HR images of $48s \times 48s$ pixels from the original training images.  Subsequently, these HR images are down-sampled to their corresponding $48 \times 48$ pixel LR counterparts using bicubic interpolation.
The dimensions of each HR image can be interpreted as a set of coordinate-patch pairs, with a total count of $(48s)^2$. From this set, a fixed number of $48^2$ pairs are selected as the training data to ensure consistency in the quantity of training data samples across different patches.

\vspace{-5pt}
\paragraph{Evaluation metrics.}
In our experiments, fidelity-oriented metrics, such as Peak Signal-to-Noise Ratio (PSNR) and Structural Similarity Index (SSIM), are reported to facilitate a fair comparison with existing methods.
However, PSNR and SSIM are known to be insufficient in reflecting perceptual quality for SR tasks. Therefore, an alternative metric, referred to as LPIPS\cite{lpips}, is employed to evaluate perceptual quality. 
Moreover, a Diversity metric, defined as the pixel value standard deviation of five samples, 
is utilized when comparing LINF with generative SR models to highlight the diversity of the SR images generated by LINF. 
\vspace{-5pt}
\paragraph{Inference temperature.}

While the flow model maps the target distribution to a standard normal distribution $\mathcal{N}(0, 1)$ during the training phase, temperature can be adjusted in the testing phase.
In the deterministic setting ($\tau\!=\!0$), the flow model operates similarly to PSNR-oriented SR models by generating the mean of the learned distribution. 
In contrast, when employing random samples with $\tau>0$, the flow model generates diverse and photo-realistic results. 
We report both deterministic and random sample outcomes to demonstrate the distinct characteristics of our flow model.

\subsection{Arbitrary-Scale SR}
\label{subsec:ArbitSR}
\input{tables/arbit_benchmark.tex}

\begin{figure*}[t]
  \centering
  \includegraphics[width=0.95\linewidth]{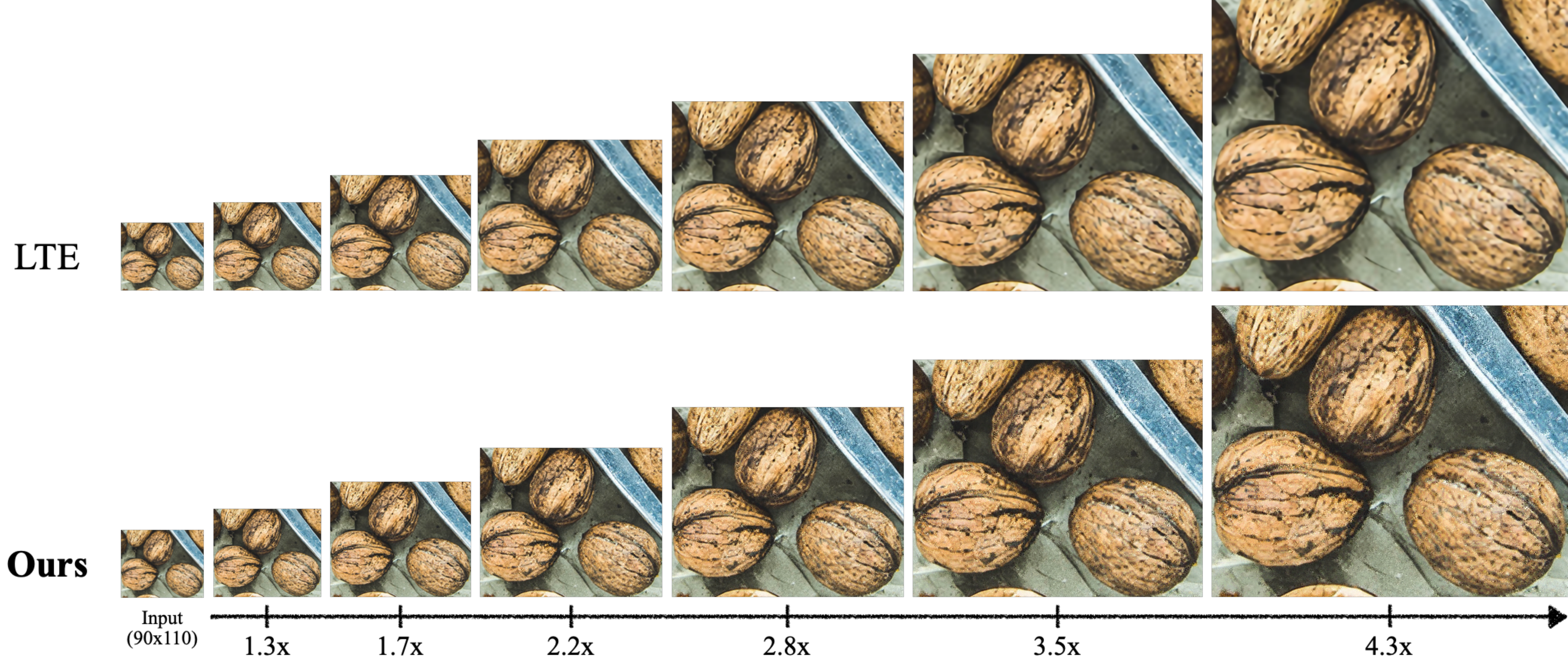}
  \caption{
    A comparison of the qualitative results evaluated by LTE~\cite{lte} and our proposed LINF for arbitrary-scale SR.
  }
  \label{fig:qualtative_arbitrary_scale}
\end{figure*}


Table~\ref{tab:exp:arbit_benchmark} presents a quantitative comparison between our LINF and the previous arbitrary-scale SR models~\cite{metasr, liif, lte}. Unlike previous arbitrary-scale SR methods, which only report PSNR, 
we take LPIPS into consideration to reflect the perceptual quality. We report results under deterministic and random sampling settings to validate the effectiveness of our model. 
In the random sample setting, we set $\tau_0$ to 0.5 for $\times2$-$\times4$ SR. As the SR scale increases, we decrease the sampling temperature to obtain more stable outputs by setting $\tau_0=0.4$ for $\times$6 SR and $\tau_0=0.2$ for $\times$8 SR. 
Our observations reveal that LINF significantly outperforms the prior methods in terms of the LPIPS metric when utilizing random sampling, indicating its ability to generate images with enhanced perceptual quality.
The qualitative results depicted in Fig.~\ref{fig:qualtative_arbitrary_scale} support the above findings, indicating that LINF can generate rich texture under arbitrary scales, while the previous PSNR-oriented method generates blurrier outcomes.
Moreover, LINF maintains competitive performance in terms of PSNR under the deterministic setting, validating that the learned distribution is centered around the average of all plausible HR images.

\subsection{Generative SR}
\label{subsec:GenerativeSR}

\begin{figure*}[t]
  \centering
  \includegraphics[width=0.95\textwidth]{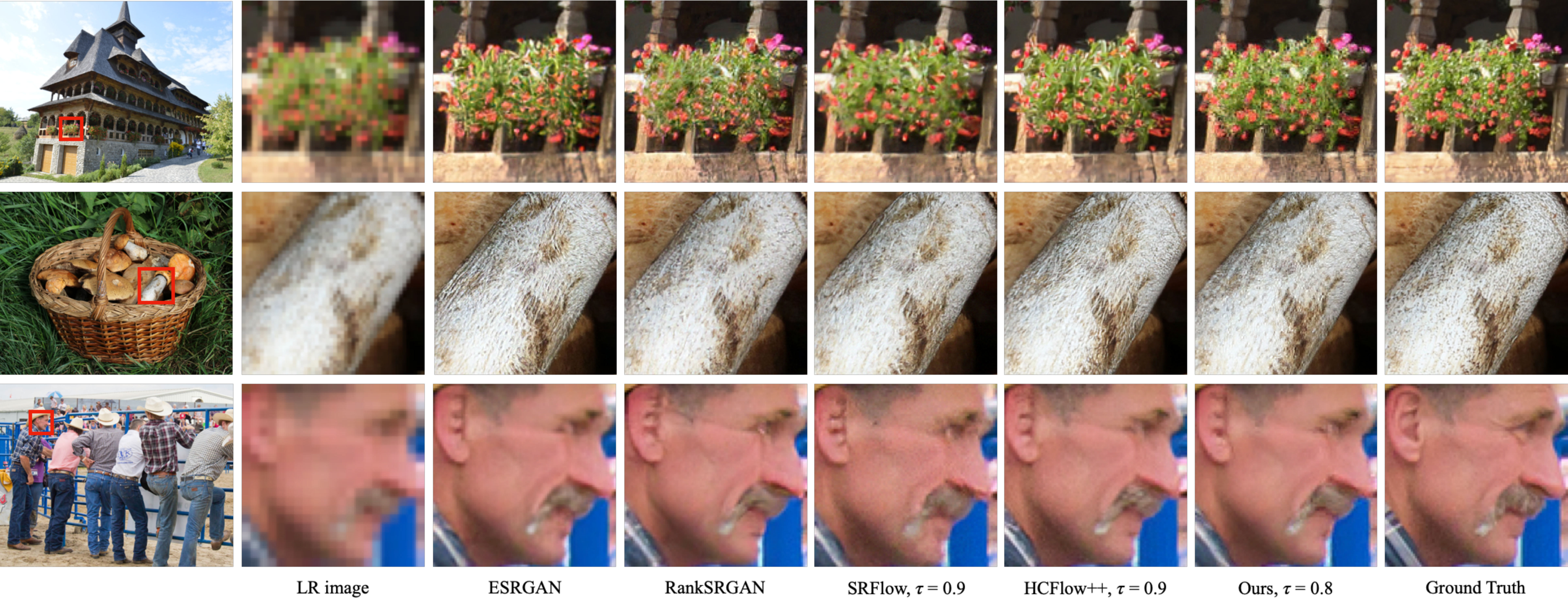}
  \caption{
    The $\times$4 SR qualitative results of generative SR methods on the DIV2K~\cite{div2k} validation set.
  }
  \label{fig:qualtative_result}
\end{figure*}

\vspace{-4pt}
\paragraph{Quantitative and qualitative results.}
We compare LINF with GAN-based~\cite{esrgan, ranksrgan}, Diffusion-based~\cite{srdiff}, AR-based\cite{larsr}, and flow-based \cite{srflow, hcflow} SR models in Table~\ref{tab:exp:generative} and Fig~\ref{fig:qualtative_result}.
HCFlow+ and HCFlow++ are two versions of HCFlow~\cite{hcflow}.  
The former employs fine-tuning with an L1 loss to enhance its PSNR performance, while the latter incorporates a VGG loss~\cite{vggloss} and an adversarial loss to improve visual quality and LPIPS scores.
In the random sampling setting,
LINF outperforms all the baselines in terms of both PSNR and LPIPS, except for SRDiff and HCFlow++. Although LINF exhibits a marginally lower PSNR than SRDiff, it significantly surpasses SRDiff in LPIPS.
Moreover, LINF outperforms HCFlow++ in PSNR with a comparable LPIPS score. These results suggest that LINF is a balanced model excelling in both PSNR and LPIPS, and are further corroborated by Fig~\ref{fig:qualtative_result}.
In the first row, SRFlow yields blurry results, while HCFlow and GAN-based models
generate over-sharpened artifacts. On the other hand, LINF generates rich textures and achieves high fidelity when compared to the ground truth image.
This evidence validates the effectiveness of LINF as a versatile and balanced model for achieving optimal performance in both PSNR and LPIPS metrics.

\input{tables/generative.tex}

\begin{figure}[t]
  \centering
  \includegraphics[width=0.95\linewidth]{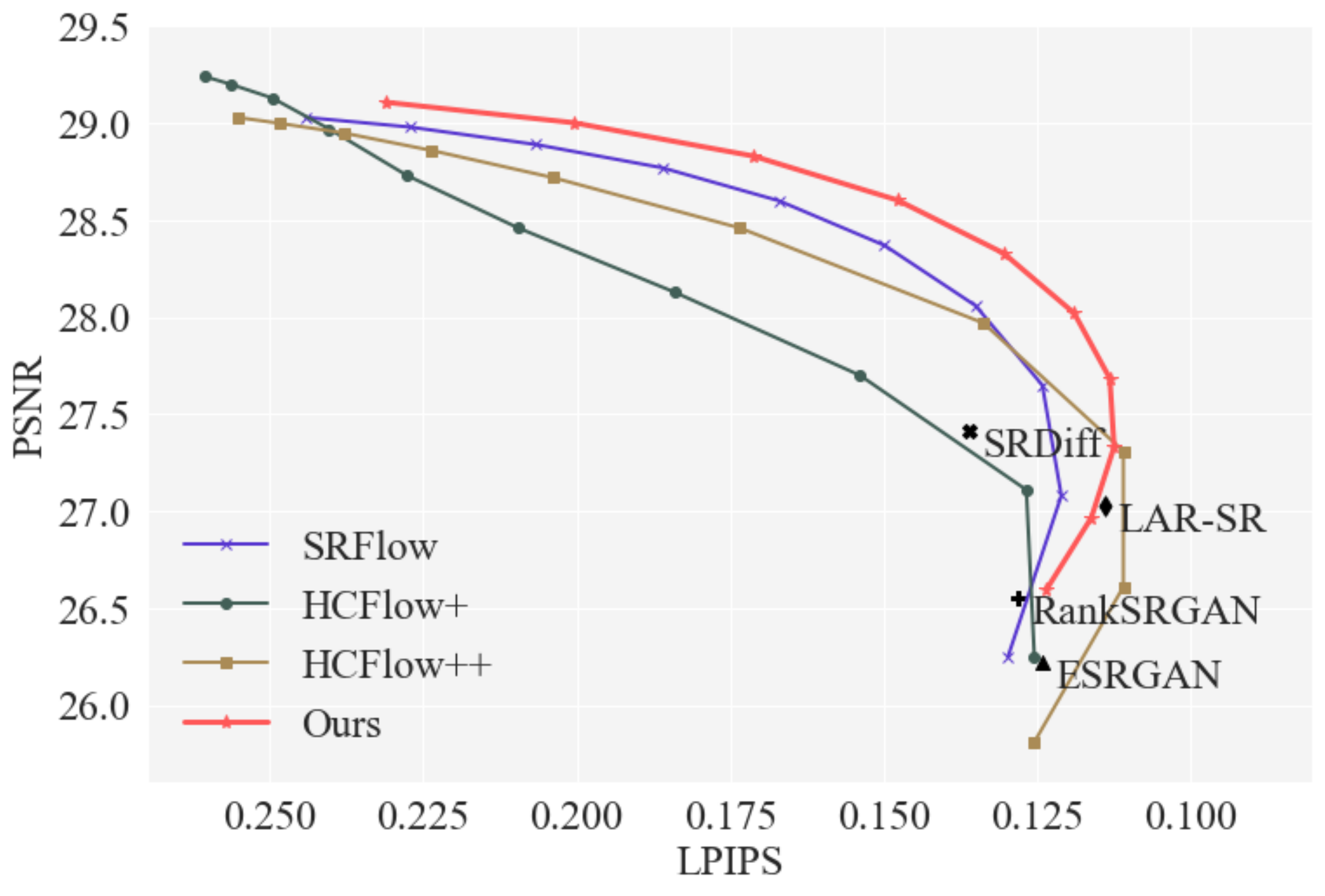}
  \caption{
  An illustration of the trade-off between PSNR and LPIPS with varying sampling temperatures $\tau$. The sampling temperature increases from the top left corner ($t=0.0$) to the bottom right corner ($t=1.0$). The x-axis is reversed for improved visualization.
  }
  \label{fig:trade_off}
\end{figure}

\begin{figure}[t]
  \centering
  \includegraphics[width=1\linewidth]{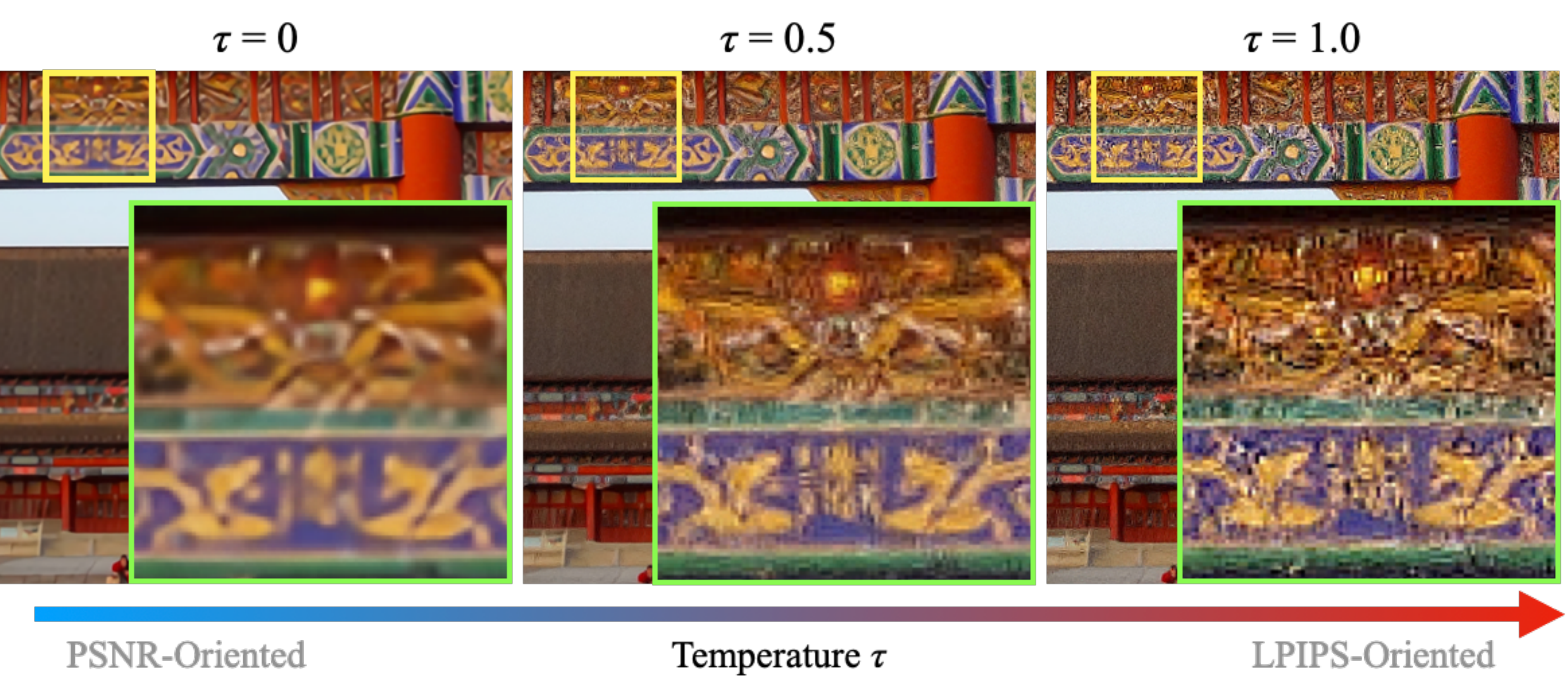}
  \caption{
    An example for depicting
    the trade-off between 
    fidelity- and perceptual-oriented results
    using different temperature $\tau$.
  }
  \label{fig:qualtative_trade_off}
\end{figure}

\vspace{-5pt}
\paragraph{Fidelity-perception trade-off.}
Since SR presents an ill-posed problem, achieving optimal fidelity (i.e., the discrepancy between reconstructed and ground truth images) and perceptual quality simultaneously presents a considerable challenge~\cite{tradeoff}. As a result, the trade-off between fidelity and perceptual quality necessitates an in-depth exploration.
By leveraging the inherent sampling property of normalizing flow, it is feasible to plot the trade-off curve between PSNR (fidelity) and LPIPS (perception) for flow-based models by adjusting temperatures, as depicted in Fig~\ref{fig:trade_off}.
This trade-off curve reveals two distinct insights. First, when the sampling temperature escalates from low to high (i.e., from the top left corner to the bottom right corner), the flow models tend to exhibit lower PSNR but improved LPIPS. However, beyond a specific temperature threshold, both PSNR and LPIPS degrade as the temperature increase.  This suggests that a higher temperature does not guarantee enhanced perceptual quality, as flow models may generate noisy artifacts. Nevertheless, through appropriate control of the sampling temperature, it is possible to select the preferred trade-off between fidelity and visual quality to produce photo-realistic images, as demonstrated in Fig~\ref{fig:qualtative_trade_off}.  Second, Fig~\ref{fig:trade_off} illustrates that the trade-off Pareto front of LINF consistently outperforms those of the prior flow-based methods except at the two extreme ends.  This reveals that given an equal PSNR, LINF exhibits superior LPIPS.  Conversely, when LPIPS values are identical, LINF demonstrates improved PSNR. This finding underscores that LINF attains a more favorable balance between PSNR and LPIPS in comparison to preceding techniques.

\vspace{-5pt}
\paragraph{Computation time.}
\input{tables/computation.tex}

To demonstrate the advantages of the proposed Fourier feature ensemble and local texture patch based generative approach in enhancing the inference speed of LINF, we compare the average inference time for a single DIV2K image with that of the contemporary generative SR models~\cite{larsr, srflow, hcflow}. As shown in Table~\ref{tab:exp:computation}, the inference time of LINF is approximately $27.2$ times faster than the autoregressive (AR)-based SR models \cite{larsr} and $2.6$ times faster than the flow-based SR models \cite{srflow, hcflow}, while concurrently achieving competitive performance in terms of the LPIPS metric.


\input{tables/Table4-5.tex}

\subsection{Ablation Study} 
\label{subsec:ablation}
\paragraph{Fourier feature ensemble.}

As discussed in Section~\ref{subsec:LINR},  LINF employs a Fourier feature ensemble mechanism to replace the local ensemble mechanism. To validate its effectiveness, we compare the two mechanisms in
%
Table~\ref{tab:exp:table4-5}. The results show that the former 
reduces the inference time by approximately $33\%$ compared to
the latter,
while maintaining a competitive performance on the SR metrics. Moreover, neglecting to scale the amplitude of the Fourier features with ensemble weights results in a slightly worse performance. This validates that scaling the amplitude of the Fourier features with ensemble weights is effective, and enables LINF to focus on the more important information. 

\vspace{-5pt}
\paragraph{Analysis of the impact of local region size.}

\begin{figure}[t]
  \centering
  \includegraphics[width=0.95\linewidth]{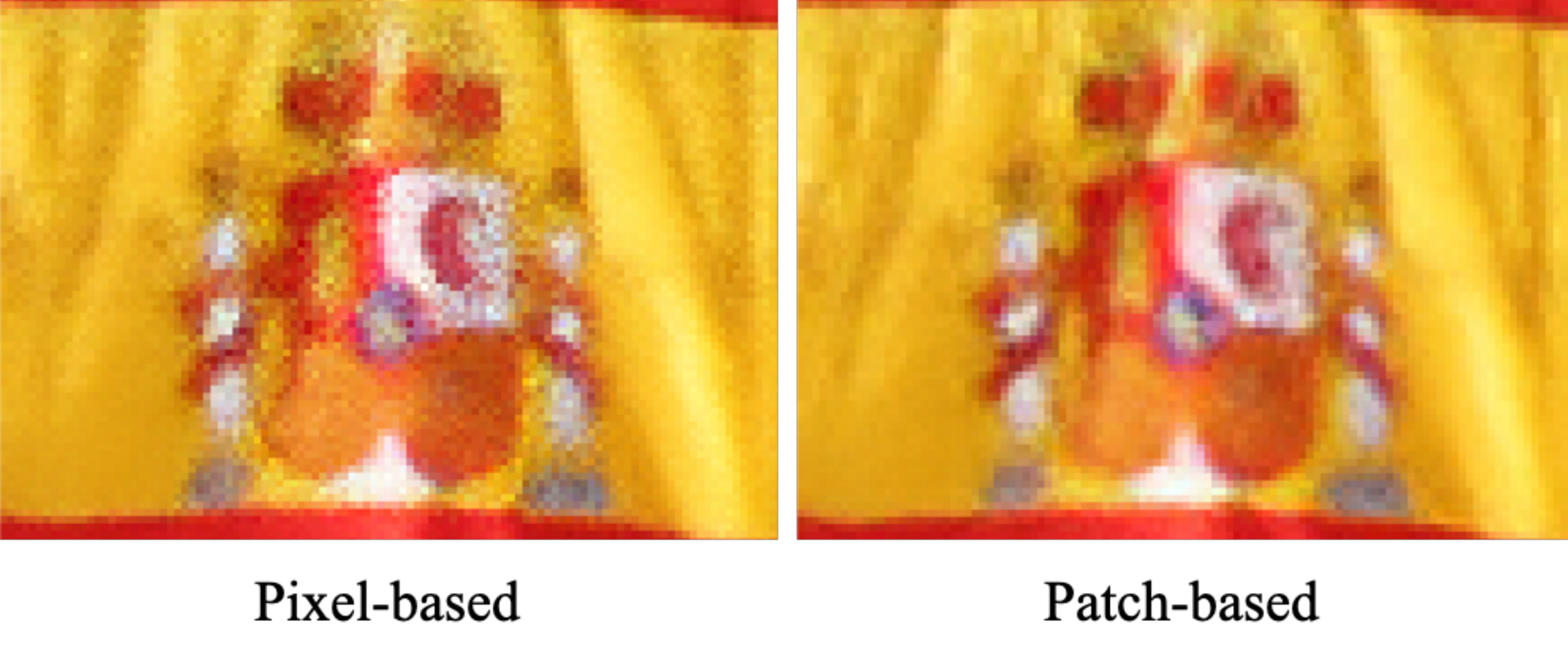}
  \caption{
    The local incoherence issue of the pixel-based method.  Note that both images are sampled with a temperature of $\tau=0.6$.
  }
  \label{fig:local_coherence}
\end{figure}
As described in Section~\ref{sec::methodology},
our proposed framework aims to learn the texture distribution of an $n \times n$ local region, where $n$ governs the region size. 
As a result, our model can be categorized as either pixel-based and patch-based by setting $n = 1$ and $n > 1$, respectively.
Table~\ref{tab:exp:table4-5} also presents a quantitative comparison between pixel-based and patch-based models.
The results reveal that a pixel-based model can generate high-fidelity images with a superior PSNR compared to a patch-based one when the temperature is set to zero. 
However, in the random sample setting, a patch-based model can generate higher perceptual quality images with a lower LPIPS. 
This phenomenon is attributed to the local-incoherent issue when sampling with pixel-based method. Specifically, pixel-wise random sampling can occasionally result in 
incoherent color, as illustrated in Fig~\ref{fig:local_coherence}. 
In contrast, a patch-based model preserves local coherency by considering the distribution of a patch, thereby achieving enhanced visual quality.
In addition, while a pixel-based model requires $H \times W$ forward passes to generate an image of shape $H \times W$, a patch-based model necessitates only $(\lceil H / n \rceil) \times (\lceil W / n \rceil)$ forward passes, yielding greater efficiency in inference, particularly for larger values of $n$.

%% file: tables/arbit_benchmark.tex

\begin{table*}[t]
\newcommand{\mytoprule}{\toprule[1.2pt]}
\centering
\setlength{\tabcolsep}{1.0em}
\footnotesize
\resizebox{1.8\columnwidth}{!}{%
    \begin{tabular}{c|c|c|c|c|c|c|c|c|c|c}
        \hline
         & \multicolumn{5}{c}{Set5} & \multicolumn{5}{|c}{Set14} \\ \cline{2-11}
         
        Method & \multicolumn{3}{c|}{In-Scale} & \multicolumn{2}{c}{Out-of-scale} & \multicolumn{3}{|c|}{In-Scale} & 
        \multicolumn{2}{c}{Out-of-scale} \\
        
        & \multicolumn{1}{c}{$\times$2} & \multicolumn{1}{c}{$\times$3} & \multicolumn{1}{c|}{$\times$4} & \multicolumn{1}{c}{$\times$6} & \multicolumn{1}{c}{$\times$8} & \multicolumn{1}{|c}{$\times$2} & \multicolumn{1}{c}{$\times$3} & \multicolumn{1}{c|}{$\times$4} & \multicolumn{1}{c}{$\times$6} & \multicolumn{1}{c}{$\times$8} \\ \hline \hline
        
        EDSR-baseline-MetaSR & 37.96 / 0.057 & 34.38 / 0.125 & 32.07 / 0.175 & 28.67 / 0.253 & 26.73 / 0.326 & 33.60 / 0.094 & 30.29 / 0.207 & 28.52 / 0.286 & 26.31 / 0.395 & 24.81 / 0.460 \\
        
        EDSR-baseline-LIIF & 37.99 / 0.056 & 34.40 / 0.124 & 32.24 / 0.173 & 28.96 / 0.248 & 26.98 / 0.307 & 33.66 / 0.093 & 30.34 / 0.205 & 28.62 / 0.284 & 26.45 / 0.390 & 24.94 / 0.449 \\ 
        
        EDSR-baseline-LTE & \textbf{\underline{38.04}} / 0.056 & 34.43 / 0.123 & 32.24 / 0.174 & \textbf{\underline{28.97}} / 0.257 & \textbf{\underline{27.04}} / 0.326 & \textbf{\underline{33.72}} / 0.092 & \textbf{\underline{30.37}} / 0.203 & \textbf{\underline{28.65}} / 0.283 & \textbf{\underline{26.50}} / 0.396 & \textbf{\underline{24.99}} / 0.463 \\
        
        \textbf{EDSR-baseline-Ours} $\tau$ = 0 & 38.00 / 0.058 & \textbf{\underline{34.45}} / 0.125 & \textbf{\underline{32.26}} / 0.176 & 28.91 / 0.251 & 26.96 / 0.312 & 33.62 / 0.096 & 30.33 / 0.207 & 28.63 / 0.286 & 26.46 / 0.395 & 24.93 / 0.457 \\
        
        \textbf{EDSR-baseline-Ours} $\tau$ = $\tau_0$ & 37.06 / \textbf{\underline{0.039}} & 33.39 / \textbf{\underline{0.067}} & 31.00 / \textbf{\underline{0.087}} & 27.88 / \textbf{\underline{0.173}} & 26.69 / \textbf{\underline{0.254}} & 32.73 / \textbf{\underline{0.071}} & 29.26 / \textbf{\underline{0.129}} & 27.54 / \textbf{\underline{0.189}} & 25.71 / \textbf{\underline{0.314}} & 24.74 / \textbf{\underline{0.396}} \\ \hline

        RDN-MetaSR & 38.22 / 0.055 & 34.65 / 0.124 & 32.40 / 0.173 & 28.99 / 0.246 & 26.93 / 0.316 & \textbf{\underline{34.15}} / 0.086 & 30.55 / 0.200 & 28.80 / 0.279 & 26.50 / 0.381 & 24.95 / 0.444 \\

        RDN-LIIF & 38.17 / 0.055 & 34.68 / 0.122 & 32.50 / {0.170} & 29.15 / {0.240} & 27.14 / {0.299} & {33.97} / 0.088 & 30.53 / {0.197} & 28.80 / 0.277 & 26.64 / {0.379} & {25.15} / {0.438} \\
        
        RDN-LTE & \textbf{\underline{38.23}} / 0.055 & \textbf{\underline{34.72}} / 0.122 & \textbf{\underline{32.61}} / 0.171 & \textbf{\underline{29.32}} / 0.253 & \textbf{\underline{27.26}} / 0.316 & 34.09 / {0.087} & \textbf{\underline{30.58}} / 0.198 & \textbf{\underline{28.88}} / 0.277 & \textbf{\underline{26.71}} / 0.389 & \textbf{\underline{25.16}} / 0.455 \\
        
        \textbf{RDN-Ours} $\tau$ = 0 & {38.21} / 0.056 & {34.71} / 0.122 & 32.50 / 0.172 & {29.21} / 0.244 & {27.23} / 0.304 & 33.91 / 0.089 & {30.56} / 0.199 & {28.83} / 0.277 & {26.65} / 0.386 & 25.14 / 0.445 \\ 
        
        \textbf{RDN-Ours} $\tau$ = $\tau_0$ & 37.36 / \textbf{\underline{0.038}} & 33.76 / \textbf{\underline{0.065}} &31.38 / \textbf{\underline{0.081}} &28.32 / \textbf{\underline{0.160}} &27.00 / \textbf{\underline{0.246}} &33.09 / \textbf{\underline{0.068}} &29.60 / \textbf{\underline{0.125}} &27.77 / \textbf{\underline{0.179}} &25.95 / \textbf{\underline{0.300}} &24.95 / \textbf{\underline{0.381}} \\ \hline
        
        SwinIR-MetaSR & 38.26 / {0.055} & 34.77 / 0.120 & 32.47 / 0.168 & 29.09 / 0.237 & 27.02 / 0.314 & 34.14 / 0.086 & 30.66 / 0.195 & 28.85 / 0.272 & 26.58 / 0.379 & 25.09 / 0.446 \\
        
        SwinIR-LIIF & {38.28} / {0.055} & {34.87} / 0.118 & 32.73 / 0.168 & {29.46} / 0.234 & {27.36} / 0.293 & 34.14 / 0.087 & {30.75} / 0.194 & {28.98} / 0.273 & 26.82 / 0.377 & {25.34} / 0.435 \\ 
        
        SwinIR-LTE & \textbf{\underline{38.33}} / {0.055} & \textbf{\underline{34.89}} / 0.120 & \textbf{\underline{32.81}} / 0.170 & \textbf{\underline{29.50}} / 0.243 & 27.35 / 0.308 & \textbf{\underline{34.25}} / 0.086 & \textbf{\underline{30.80}} / 0.194 & \textbf{\underline{29.06}} / 0.270 & \textbf{\underline{26.86}} / 0.382 & \textbf{\underline{25.42}} / 0.449 \\
        
        \textbf{SwinIR-Ours} $\tau$ = 0 & {38.28} / 0.056 & 34.85 / 0.121 & {32.74} / 0.170 & 29.40 / 0.238 & \textbf{\underline{27.45}} / 0.294 & 34.13 / 0.087 & 30.71 / 0.195 & 28.95 / 0.273 & {26.84} / 0.376 & 25.30 / 0.436 \\
        
        \textbf{SwinIR-Ours} $\tau$ = $\tau_0$ & 37.49 / \textbf{\underline{0.038}} & 33.94 / \textbf{\underline{0.066}} & 31.70 / \textbf{\underline{0.084}} & 28.49 / \textbf{\underline{0.153}} & 27.19 / \textbf{\underline{0.236}} & 33.38 / \textbf{\underline{0.067}} & 29.84 / \textbf{\underline{0.127}} & 27.98 / \textbf{\underline{0.176}} & 26.15 / \textbf{\underline{0.286}} & 25.09 / \textbf{\underline{0.370}} \\ \hline \hline
        
        & \multicolumn{5}{c}{B100} & \multicolumn{5}{|c}{Urban100} \\ \cline{2-11}
         
        Method & \multicolumn{3}{c|}{In-Scale} & \multicolumn{2}{c}{Out-of-scale} & \multicolumn{3}{|c|}{In-Scale} & 
        \multicolumn{2}{c}{Out-of-scale} \\
        
        & \multicolumn{1}{c}{$\times$2} & \multicolumn{1}{c}{$\times$3} & \multicolumn{1}{c|}{$\times$4} & \multicolumn{1}{c}{$\times$6} & \multicolumn{1}{c}{$\times$8} & \multicolumn{1}{|c}{$\times$2} & \multicolumn{1}{c}{$\times$3} & \multicolumn{1}{c|}{$\times$4} & \multicolumn{1}{c}{$\times$6} & \multicolumn{1}{c}{$\times$8} \\ \hline \hline
        
        EDSR-baseline-MetaSR & 32.17 / 0.147 & 29.09 / 0.285 & 27.55 / 0.376 & 25.76 / 0.492 & 24.70 / 0.565 & 32.10 / 0.065 & 28.12 / 0.157 & 25.96 / 0.233 & 23.59 / 0.352 & 22.30 / 0.446 \\
        
        EDSR-baseline-LIIF & 32.17 / 0.147 & 29.10 / 0.282 & 27.60 / 0.372 & 25.84 / 0.486 & 24.79 / 0.556 & 32.15 / 0.064 & 28.22 / 0.155 & 26.15 / 0.228 & 23.79 / 0.338 & 22.45 / 0.422 \\
        
        EDSR-baseline-LTE & \textbf{\underline{32.21}} / 0.146 & \textbf{\underline{29.14}} / 0.280 & \textbf{\underline{27.62}} / 0.371 & \textbf{\underline{25.87}} / 0.495 & \textbf{\underline{24.82}} / 0.570 & \textbf{\underline{32.29}} / 0.063 & \textbf{\underline{28.32}} / 0.152 & \textbf{\underline{26.24}} / 0.224 & \textbf{\underline{23.85}} / 0.345 & \textbf{\underline{22.53}} / 0.436 \\
        
        \textbf{EDSR-baseline-Ours} $\tau$ = 0 & 32.16 / 0.151 & 29.12 / 0.286 & 27.61 / 0.374 & 25.85 / 0.492 & 24.80 / 0.563 & 32.11 / 0.066 & 28.21 / 0.157 & 26.15 / 0.232 & 23.79 / 0.344 & 22.45 / 0.431 \\
        
        \textbf{EDSR-baseline-Ours} $\tau$ = $\tau_0$ & 31.39 / \textbf{\underline{0.114}} & 28.21 / \textbf{\underline{0.174}} & 26.62 / \textbf{\underline{0.238}} & 25.21 / \textbf{\underline{0.382}} & 24.64 / \textbf{\underline{0.486}} & 31.22 / \textbf{\underline{0.052}} & 27.26 / \textbf{\underline{0.115}} & 25.15 / \textbf{\underline{0.184}} & 23.11 / \textbf{\underline{0.331}} & 22.27 / \textbf{\underline{0.415}} \\ \hline

        RDN-MetaSR & 32.34 / 0.143 & 29.26 / 0.282 & 27.71 / 0.369 & 25.89 / 0.477 & 24.82 / 0.549 & 32.96 / 0.055 & 28.87 / 0.140 & 26.60 / 0.211 & 24.00 / 0.317 & 22.59 / 0.408 \\
        
        RDN-LIIF & {32.32}/ 0.145 & 29.26 / 0.278 & 27.74 / 0.365 & 25.98 / {0.475} & 24.91 / {0.544} & {32.87} / 0.057 & {28.82} / 0.139 & 26.68 / 0.209 & {24.20} / {0.312} & {22.79} / 0.392 \\ 
        
        RDN-LTE & \textbf{\underline{32.36}} / {0.142} & \textbf{\underline{29.30}} / {0.275} & \textbf{\underline{27.77}} / {0.363} & \textbf{\underline{26.01}} / 0.485 & \textbf{\underline{24.95}} / 0.561 & \textbf{\underline{33.04}} / {0.055} & \textbf{\underline{28.97}} / {0.138} & \textbf{\underline{26.81}} / {0.206} & \textbf{\underline{24.28}} / 0.324 & \textbf{\underline{22.88}} / 0.412 \\
        
        \textbf{RDN-Ours} $\tau$ = 0 & 32.31 / 0.145 & 29.26 / 0.279 & {27.75} / 0.366 & {26.00} / 0.482 & {24.93} / 0.555 & 32.86 / 0.057 & 28.81 / 0.140 & {26.69} / 0.210 & 24.19 / 0.317 & 22.77 / {0.403} \\
        
        \textbf{RDN-Ours} $\tau$ = $\tau_0$ & 31.60 / \textbf{\underline{0.111}} & 28.46 / \textbf{\underline{0.173}} & 26.84 / \textbf{\underline{0.229}} & 25.40 / \textbf{\underline{0.365}} & 24.78 / \textbf{\underline{0.470}} & 32.06 / \textbf{\underline{0.044}} & 27.97 / \textbf{\underline{0.099}} & 25.79 / \textbf{\underline{0.155}} & 23.55 / \textbf{\underline{0.288}} & 22.60 / \textbf{\underline{0.376}} \\ \hline

        SwinIR-MetaSR & 32.39 / 0.141 & 29.31 / 0.280 & 27.75 / 0.365 & 25.94 / 0.472 & 24.87 / 0.549 & 33.29 / 0.052 & 29.12 / 0.132 & 26.76 / 0.200 & 24.16 / 0.315 & 22.75 / 0.403 \\
        
        SwinIR-LIIF & 32.39 / 0.143 & 29.34 / 0.277 & {27.84} / 0.362 & 26.07 / {0.469} & 25.01 / {0.539} & {33.36} / 0.054 & {29.33} / 0.133 & {27.15} / 0.201 & {24.59} / 0.299 & {23.14} / 0.377 \\
        
        SwinIR-LTE & \textbf{\underline{32.44}} / {0.139} & \textbf{\underline{29.39}} / {0.270} & \textbf{\underline{27.86}} / {0.357} & \textbf{\underline{26.09}} / 0.476 & \textbf{\underline{25.03}} / 0.553 & \textbf{\underline{33.50}} / 0.052 & \textbf{\underline{29.41}} / {0.130} & \textbf{\underline{27.24}} / {0.194} & \textbf{\underline{24.62}} / 0.309 & \textbf{\underline{23.17}} / 0.396 \\
        
        \textbf{SwinIR-Ours} $\tau$ = 0 & 32.39 / 0.142 & 29.34 / 0.273 & 27.83 / 0.361 & \textbf{\underline{26.09}} / 0.470 & {25.02} / 0.542 & 33.27 / 0.053 & 29.23 / 0.133 & 27.06 / 0.200 & 24.54 / 0.299 & 23.08 / {0.379} \\
        
        \textbf{SwinIR-Ours} $\tau$ = $\tau_0$ & 31.72 / \textbf{\underline{0.110}} & 28.55 / \textbf{\underline{0.170}} & 26.96 / \textbf{\underline{0.223}} & 25.46 / \textbf{\underline{0.352}} & 24.85 / \textbf{\underline{0.457}} & 32.49 / \textbf{\underline{0.042}} & 28.39 / \textbf{\underline{0.093}} & 26.16 / \textbf{\underline{0.144}} & 23.78 / \textbf{\underline{0.267}} & 22.85 / \textbf{\underline{0.351}} \\ \hline

    \end{tabular}
    
}
\caption{The arbitrary-scale SR results of the baselines and LINF (denoted as ``\textit{Ours}") evaluated on the widely used SR benchmark datasets~\cite{set5, set14, b100, urban100}. Note that PSNR is evaluated on the Y channel of the YCbCr space. The best results are denoted in bold and underlined.}
\label{tab:exp:arbit_benchmark}
\end{table*}

%% file: tables/generative.tex
\begin{table}[t]
\renewcommand{\arraystretch}{1.3}
\newcommand{\mytoprule}{\toprule[1.2pt]}
\centering
\footnotesize
\resizebox{0.9\columnwidth}{!}{%
    \begin{tabular}{l|c|c|c|c}
        \hline
        Method & PSNR$\uparrow$ & SSIM$\uparrow$ & LPIPS$\downarrow$ & Diversity$\uparrow$  \\
        \hline
        \hline
        ESRGAN\cite{esrgan} & 26.22 & 0.75 & 0.124 & 0 \\
        RankSRGAN\cite{ranksrgan}& 26.55 & 0.75 & 0.128 & 0 \\
        SRDiff\cite{srdiff} & \textcolor{red}{27.41} & \textcolor{red}{0.79} & 0.136 & 6.1 \\
        LAR-SR\cite{larsr} & 27.03 & \textcolor{blue}{0.77} & 0.114 & - \\
        SRFlow $\tau=0.9$\cite{srflow} & 27.08 & 0.76 & 0.121 & 5.6 \\
        HCFlow+ $\tau=0.9$\cite{hcflow} & 27.11 & 0.76 & 0.127 & 4.7 \\
        HCFlow++ $\tau=0.9$\cite{hcflow} & 26.61 & 0.74 & \textcolor{red}{0.111} & 5.4 \\
        \textbf{Ours} $\tau=0.8$ & \textcolor{blue}{27.33} & 0.76 & \textcolor{blue}{0.112} & 5.1 \\ 
        \hline
        SRFlow $\tau=0$\cite{srflow} & 29.05 & \textcolor{red}{0.83} & \textcolor{blue}{0.251} & 0 \\
        HCFlow+ $\tau=0$\cite{hcflow} & \textcolor{red}{29.25} & \textcolor{red}{0.83} & 0.262 & 0 \\
        HCFlow++ $\tau=0$\cite{hcflow} & 29.04 & 0.82 & 0.258 & 0 \\
        \textbf{Ours} $\tau=0$ & \textcolor{blue}{29.14} & \textcolor{red}{0.83} & \textcolor{red}{0.248} & 0 \\
        \hline
    \end{tabular}
}
\caption{The $\times$4 SR results on the DIV2K~\cite{div2k} validation set. Note that  PSNR and SSIM are evaluated on the RGB space. The best and second best results are marked in \textcolor{red}{red} and \textcolor{blue}{blue}, respectively.}
\label{tab:exp:generative}
\end{table}

%% file: tables/computation.tex
\begin{table}[t]
\renewcommand{\arraystretch}{1.3}
\newcommand{\mytoprule}{\toprule[1.2pt]}
\centering
\footnotesize
\resizebox{.95\columnwidth}{!}{
    \begin{tabular}{l|c|c|c}
        \hline
        Method  & LPIPS $\downarrow$ & Time (s)$\downarrow$ & \#Param \\
        \hline
        \hline
        LAR-SR~\cite{larsr} & 0.114 & 14.70 & 62.1M \\
        SRFlow $\tau=0.9$~\cite{srflow} & 0.121 & 1.43 & 39.5M \\
        HCFlow++ $\tau=0.9$~\cite{hcflow} & \textbf{\underline{0.111}} & 1.46 & 23.2M\\
        Ours $\tau=0.8$ & 0.112 & \textbf{\underline{0.54}} & \textbf{\underline{17.5M}} \\
        \hline
    \end{tabular}
}
\caption{The average $\times$4 SR inference time of a single DIV2K~\cite{div2k} image. The computation time is evaluated on an NVIDIA Tesla V100. The best results are denoted in bold and underlined.}
\label{tab:exp:computation}
\end{table}

%% file: tables/Table4-5.tex
\begin{table}[t]
\renewcommand{\arraystretch}{1.3}
\newcommand{\mytoprule}{\toprule[1.2pt]}
\centering
\footnotesize
\resizebox{.95\columnwidth}{!}{%
    \begin{tabular}{l|c|c|c|c}
        \hline
        Method & PSNR$\uparrow$ & SSIM$\uparrow$ & LPIPS$\downarrow$ & Time (s)$\downarrow$ \\
        \hline
        \hline
        Local ensemble & \textbf{\underline{29.04}} & 0.82 & 0.270 & 2.16 \\
        Fourier ensemble & \textbf{\underline{29.04}} & 0.82 & 0.270 & 1.44 \\
        Fourier ensemble (-W) & 29.03 & 0.82 & 0.271 & 1.39 \\
        \hline
        Fourier ensemble (+P) $\tau = 0$ & 28.85 & 0.82 & 0.273 & \textbf{\multirow{2}{*}{\underline{0.33}}} \\
        Fourier ensemble (+P) $\tau = 0.6$ & 27.43 & 0.77 & \textbf{\underline{0.158}} \\
        \hline
    \end{tabular}
}
\caption{The $\times$4 SR results on the DIV2K\cite{div2k} validation set. EDSR-baseline~\cite{edsr} is used as the encoder, -W refers to removing the amplitude scaling, and +P indicates the usage of 3$\times$3 patch-based model. The computation time is evaluated on an NVIDIA TITAN X. The best results are denoted in bold and underlined.}
\label{tab:exp:table4-5}
\end{table}

%% file: sections/conclusions.tex
In this paper, we introduced a novel framework called LINF for arbitrary-scale SR. To the best of our knowledge, LINF is the first approach to employ normalizing flow for arbitrary-scale SR. Specifically, we formulated SR as a problem of learning the distributions of local texture patches. We utilized coordinate conditional normalizing flow to learn the distribution and a local implicit module to generate conditional signals. Through our quantitative and qualitative experiments, we demonstrated that LINF can produce photo-realistic high-resolution images at arbitrary upscaling scales while achieving the optimal balance between fidelity and perceptual quality among all methods.


%% file: sections/acknowledgment.tex
\section*{Acknowledgements}
The authors gratefully acknowledge the support from the National Science and Technology Council (NSTC) in Taiwan under grant numbers MOST 111-2223-E-007-004-MY3 and MOST 111-2628-E-007-010, as well as the financial support from MediaTek Inc., Taiwan. The authors would also like to express their appreciation for the donation of the GPUs from NVIDIA Corporation and NVIDIA AI Technology Center (NVAITC) used in this work. Furthermore, the authors extend their gratitude to the National Center for High-Performance Computing (NCHC) for providing the necessary computational and storage resources.